\documentclass[letterpaper,10pt,conference]{ieeeconf}
\IEEEoverridecommandlockouts
\usepackage[T1]{fontenc}
\usepackage{times}
\usepackage{amsmath,amssymb}
\usepackage{graphicx}
\usepackage{booktabs,array,multirow}
\usepackage{cite}
\usepackage{xcolor}
\usepackage[hidelinks]{hyperref}
\usepackage{caption}
\newif\ifanonymous
\newif\ifdraftnotes
\draftnotestrue
\newcommand{\ours}{CHOP}

\newcommand{\RMValBT}{0.3161}
\newcommand{\RMValAcc}{87.66}
\newcommand{\RMTestBT}{0.4525}
\newcommand{\RMTestAcc}{83.01}
\newcommand{\GlobalValBT}{0.3367}

\newcommand{\RewardImage}{fig/Comps.jpg}
\newif\ifrewardfigureupdated
\rewardfigureupdatedfalse

\title{\LARGE \bf
\ours{}: Counterfactual Human Preference Labels Improve Obstacle Avoidance in Visuomotor Navigation Policies
\vspace{-8pt}
}
\vspace{-4.5em}
\ifanonymous
\author{Anonymous authors}
\else
\author{Gershom Seneviratne, Jianyu An, Vaibhav Shende, Sahire Ellahy, Yaxita Amin, Kondapi Manasanjani, \\ Samarth Chopra,
Jonathan Deepak Kannan, and Dinesh Manocha\\
University of Maryland, College Park}
\fi
\hypersetup{pdftitle={CHOP: Learning Navigation Rewards from Counterfactual Human Preferences},pdfauthor={},pdfsubject={ICRA manuscript draft}}

\IEEEaftertitletext{\vspace{-3ex}
\begin{minipage}{\textwidth}
\centering
\includegraphics[width=\textwidth]{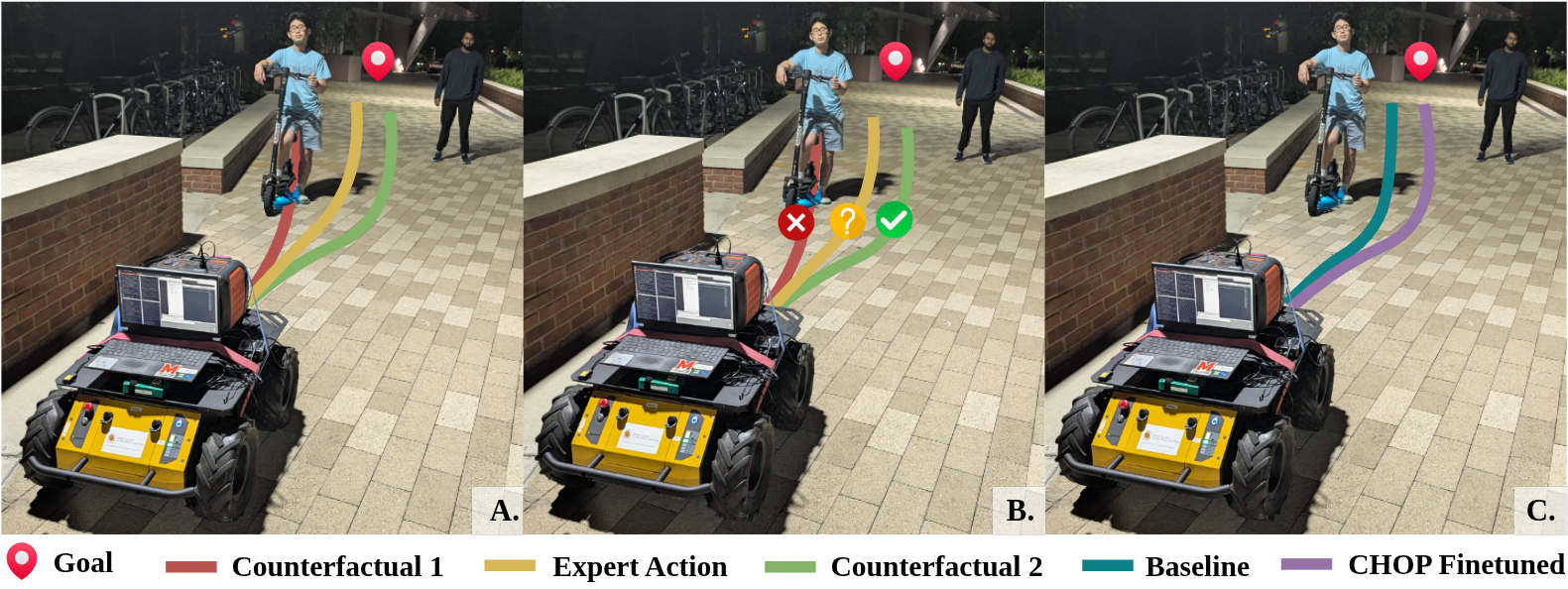}
\vspace{-1em}
\captionsetup{type=figure}
\captionof{figure}{\small \textbf{\ours{}:} Counterfactual human preferences for obstacle avoidance and planning enable alignment of visuomotor navigation policies with human preference in ambiguous environments.
\textbf{(A)} Given a single visual observation and goal, multiple distinct navigation trajectories may exist.
\textbf{(B)} Human annotators express preferences ($\checkmark$ $>$ ? $>$ $\times$)
over these counterfactual alternatives, capturing safety and contextual cues not explicitly observable, sometimes picking trajectories which are better than the operator action sequence.
\textbf{(C)} \ours{} uses these preferences to adapt visuomotor navigation policies through supervised and reward-based learning, whereas the standard SCAND-finetuned baselines trained on the recorded trajectory may select a suboptimal path. Our method shows improvements in obstacle clearance, success rate, and average collision rate in real-world scenarios.
}
\label{fig:concept}
\end{minipage}
\vspace{-1ex}

\par\bigskip
}

\begin{document}
\maketitle
\thispagestyle{empty}
\pagestyle{empty}
\begin{abstract}
Visuomotor navigation policies have shown strong perception--action coupling for embodied agents, yet they often struggle with safe navigation and obstacle avoidance in complex real-world environments. We introduce \textbf{CHOP}, an approach that leverages \textbf{Counterfactual Human Preference Labels} to improve the safety and obstacle-avoidance behavior of visuomotor navigation policies. For each visual observation, the robot's executed trajectory is considered alongside a set of counterfactual alternatives representing other possible navigation decisions under the same scene. Human annotators provide pairwise preferences over counterfactual trajectories under the same visual observation, yielding over \textbf{1.1 million comparisons}. We use these preferences in two ways: \textbf{CHOP-SFT} imitates the most preferred trajectory, while \textbf{CHOP-RFT} learns a trajectory-conditioned reward from the pairwise preferences and directly optimizes the policy using this reward. On SCAND, \textbf{CHOP-SFT} reduces near-collision events by 49.7\%, decreases deviation from human-preferred trajectories by 45.1\%, and increases obstacle clearance by 19.8\%, while \textbf{CHOP-RFT} improves these metrics by \textbf{53.1\%}, \textbf{53.3\%}, and \textbf{23.3\%}  respectively, relative to SCAND-finetuned baselines. In real-world quadruped navigation, \textbf{CHOP-SFT} improves average goal success by 28.9 percentage points, increases minimum obstacle clearance by 6.8\%, reduces collision/intervention events by 46.4\%, and improves normalized path completion by 38.7\%, while \textbf{CHOP-RFT} improves these metrics by \textbf{35.6 percentage points}, \textbf{16.0\%}, \textbf{80.1\%}, and \textbf{42.9\%}, respectively. Our results demonstrate that counterfactual human preferences provide an effective supervision signal for safety-aware visuomotor navigation through both supervised and reward-based adaptation.
\end{abstract}

\begin{figure*}[t]
\centering
\includegraphics[width=\textwidth]{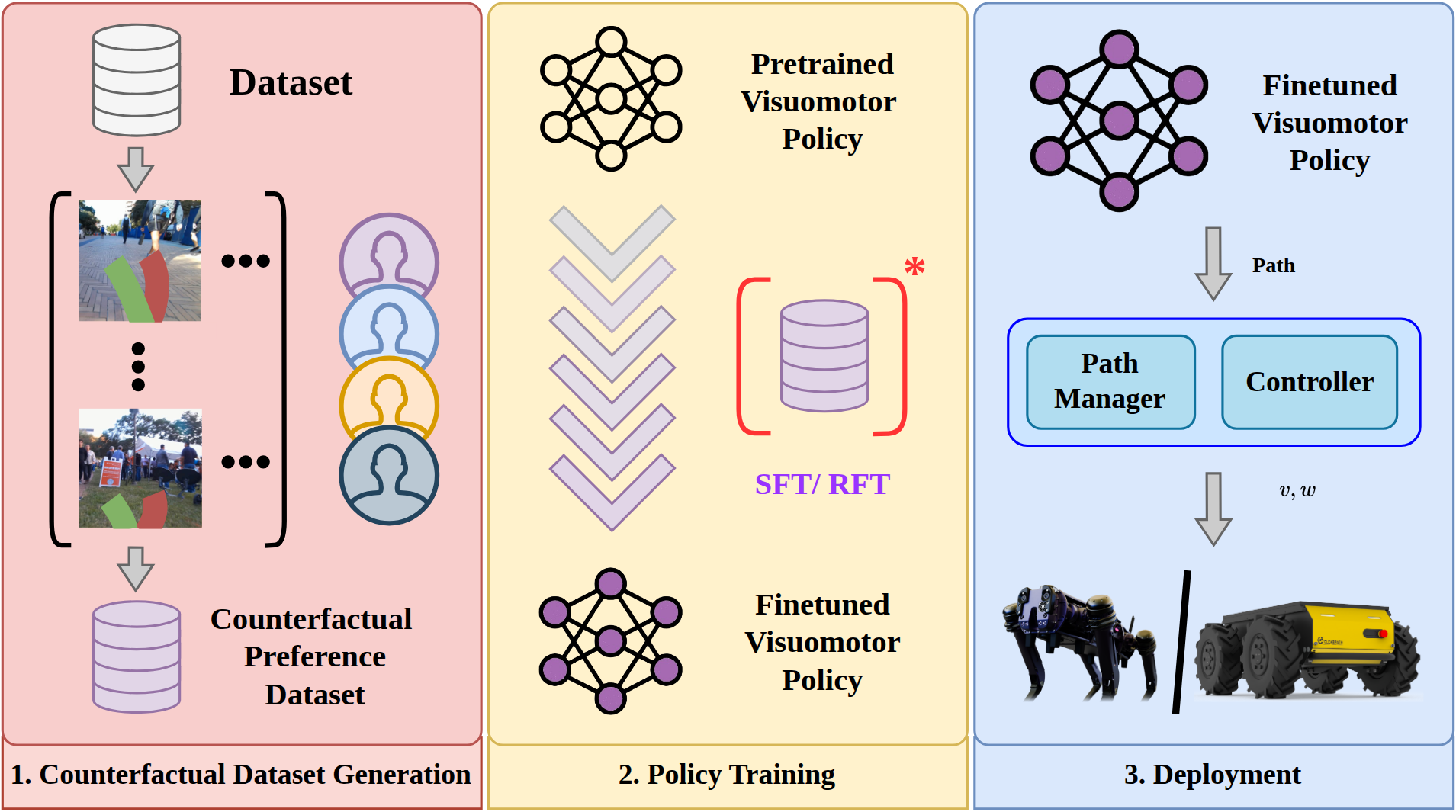}
\caption{\small{
Overview of \ours{}.
\textbf{(1) Counterfactual dataset generation:}
Given a navigation dataset containing egocentric images and robot-executed trajectories, we generate multiple counterfactual trajectory alternatives for each observation. Human annotators provide pairwise preference labels over these alternatives, resulting in multiple comparisons per observation. The resulting annotations form the CHOP counterfactual preference dataset.
\textbf{(2) Policy adaptation:}
We study two complementary ways of using the same pairwise preference supervision. In \textbf{CHOP-SFT}, the trajectory with the most pairwise wins for each observation (\textcolor{red}{\textbf{*}}) is selected as the target for supervised fine-tuning of pretrained visuomotor navigation policies. In \textbf{CHOP-RFT}, the full set of pairwise comparisons is used to learn a trajectory-conditioned visual reward model, which directly guides policy adaptation while regularizing deviation from the pretrained policy.
\textbf{(3) Deployment:}
The resulting CHOP-adapted policies are deployed on real robot platforms, where predicted trajectories are executed by a downstream controller. The learned reward model is used only during CHOP-RFT training and is not required at deployment.
}}
\label{fig:chop_arch}
\vspace{-3ex}
\end{figure*}

\section{Introduction}
\label{sec:intro}

For robots to be deployed at scale, navigation systems must be inexpensive, reliable, and robust across diverse environments. While LiDAR-based and stereo vision--based approaches can provide accurate geometric information, they often require additional sensing hardware or can become unreliable in challenging unstructured settings~\cite{hwang2025guidenav, sanket2021morpheyes}. As a result, there has been growing interest in vision-based navigation, which relies primarily on onboard cameras and offers improved scalability and deployability. Recent advances in end-to-end visuomotor navigation, including architectures that support multimodal goals and large pretrained models with broad visual and semantic knowledge, have led to substantial progress in visual navigation~\cite{shah2023vint, shah2022gnm, shah2023lm, hirose2025omnivla}. Despite this progress, visuomotor navigation policies can still exhibit risky behavior, such as approaching obstacles too closely or failing to account for dynamic agents~\cite{zhang2025creste, wang2025openbench}.

Most existing policies are trained to imitate trajectories collected from large-scale datasets. However, these trajectories are frequently obtained through teleoperation and can reflect joystick noise, execution lag, operator bias, and locally suboptimal decisions. Consequently, the demonstrated trajectory represents only one feasible action sequence for a given observation and may not be the safest or most appropriate among the available alternatives. Counterfactual trajectory supervision provides a natural way to address this limitation by considering alternative actions under the same perceptual conditions. Humans can compare such alternatives and favor trajectories that better balance obstacle clearance, safety, social interaction, and goal progress.

In this paper, we propose \textbf{CHOP} (\textbf{C}ounterfactual \textbf{H}uman preferences for \textbf{O}bstacle avoidance and \textbf{P}lanning), which collects human pairwise preferences over counterfactual navigation trajectories and uses them to adapt visuomotor policies. We study two complementary approaches: \textbf{CHOP-SFT}, which identifies the most preferred trajectory and uses it as a supervised fine-tuning target, and \textbf{CHOP-RFT}, which learns a trajectory-conditioned visual reward directly from the pairwise comparisons and uses this reward for policy adaptation.

\textbf{Main contributions:}
\begin{itemize}
    \item \textbf{Counterfactual human preference supervision:}
    We introduce a pipeline to collect pairwise human preferences over alternative navigation trajectories, providing supervision beyond imitation of a single demonstrated trajectory.

    \item \textbf{Large-scale preference dataset:}
    We augment \textbf{SCAND}~\cite{karnan2022socially} with over one million pairwise human preference labels across multiple candidate trajectories per scene.

    \item \textbf{Supervised and reward-based policy adaptation:}
    We study both \textbf{CHOP-SFT} and \textbf{CHOP-RFT} across GNM, ViNT, and OmniVLA, evaluating safety, preference alignment, and goal-directed navigation on held-out SCAND data.

    \item \textbf{Real-world validation:}
    We evaluate CHOP-aligned policies on a Ghost Robotics Vision60 quadruped in cluttered environments with static and dynamic obstacles.
\end{itemize}

To the best of our knowledge, CHOP is among the first large-scale studies to use human preferences over explicit counterfactual navigation trajectories for both supervised and reward-based adaptation of end-to-end visuomotor policies.
\section{Related Work}

\nocite{tung2026counterfactualreasoningenvironmentdesign}
\subsection{Vision-Based Navigation}

Classical navigation methods such as the Dynamic Window Approach (DWA)~\cite{fox2002dynamic} rely on hand-crafted geometric objectives and depth-based sensing for collision avoidance. Vision-based navigation instead learns navigation behavior directly from visual observations, with early work focusing on free-space and semantic understanding~\cite{hirose2018gonet, fan2020sne, guan2022ga,guerrier2026visionfoundationmodelsnavigate} and later approaches incorporating topological memory~\cite{savinov2018semi, chaplot2020neural, taniguchi2021pose}. Models such as GNM~\cite{shah2022gnm} and ViNT~\cite{shah2023vint} learn transferable goal-conditioned navigation policies from heterogeneous robot datasets, enabling generalization across environments and embodiments.

Recent work has also incorporated large vision-language models and VLA architectures into embodied navigation~\cite{song2024vlm, mu2023embodiedgpt, weerakoon2025behav, hirose2025omnivla, castro2025vamos, cheng2024navila,szvoren2026greengoredno,guerrier2026visionfoundationmodelsnavigate}. These methods primarily improve semantic understanding, instruction following, or multimodal goal conditioning. In contrast, \ours{} focuses on the supervision used to adapt visuomotor policies: rather than treating a single demonstrated trajectory as the only target, it leverages human preferences over counterfactual trajectory alternatives to improve safety-aware navigation behavior.

\subsection{Human Preference Alignment and Reward Learning}

Human preference learning commonly uses pairwise comparisons to learn reward functions or adapt policies toward preferred behavior~\cite{christiano2017deep,Wirth2017ASO,ouyang2022training, tung2026counterfactualreasoningenvironmentdesign}. In robotics, DAgger~\cite{ross2011reduction} improves imitation through expert corrections at learner-visited states, while HALO~\cite{seneviratne2025halo} learns visual navigation rewards from human-ranked actions. CREStE~\cite{zhang2025creste} uses counterfactual inverse reinforcement learning to learn navigation costs from expert and counterfactual trajectories within a modular navigation framework. In contrast, \ours{} collects pairwise human preferences over explicit trajectory alternatives, including the recorded trajectory itself, and studies two uses of this supervision: winner-based supervised fine-tuning and trajectory-conditioned reward-based adaptation of end-to-end visuomotor policies.

Our trajectory-conditioned reward model draws on spatially conditioned attention mechanisms~\cite{carion2020end,zhu2020deformable}, conditioning image attention on the candidate trajectory.



\section{Methodology}
\label{sec:methodology}

Our goal is to improve the safety of visuomotor navigation policies using counterfactual human preference supervision. Standard navigation datasets provide only the trajectory executed for a given observation, even though alternative trajectories under the same perceptual conditions may be safer or more appropriate. To address this limitation, \ours{} augments the SCAND dataset~\cite{karnan2022socially} with counterfactual trajectory alternatives and pairwise human preferences over them. As illustrated in Fig.~\ref{fig:chop_arch}, we use this preference supervision in two complementary ways: \textbf{CHOP-SFT} fine-tunes a policy toward the most preferred trajectory, while \textbf{CHOP-RFT} learns a trajectory-conditioned reward from the pairwise comparisons and uses it for policy adaptation. We describe the counterfactual data generation, reward learning, and policy adaptation below.

\subsection{Counterfactual Preference Dataset Generation}
\label{sec:dataset}

\begin{figure*}[t]
      \centering
      \includegraphics[width=\textwidth]{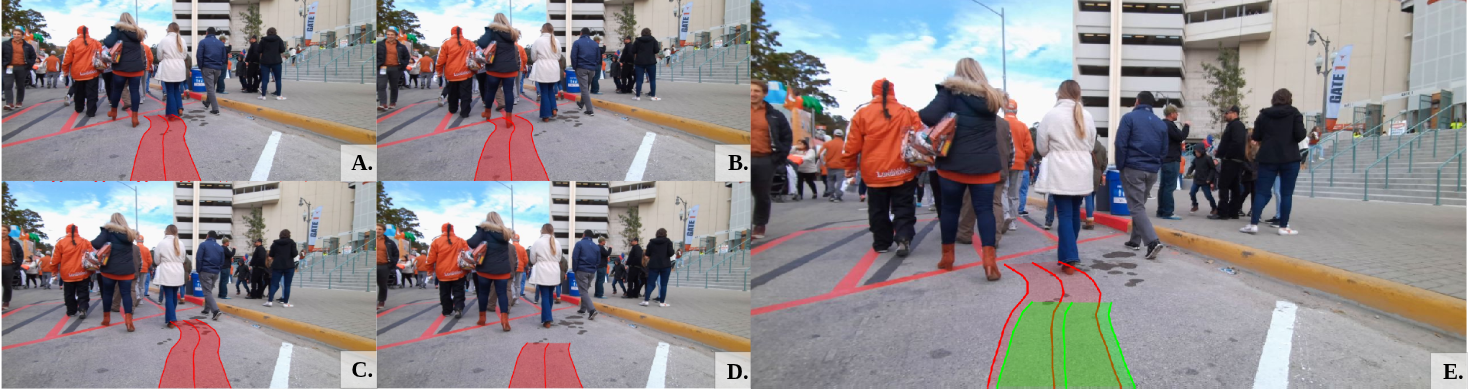}
      \caption{\small{
      Counterfactual trajectory generation and human preference annotation under identical egocentric observations.
      \textbf{A}: Trajectory recorded in the dataset.
      \textbf{B--C}: Structured counterfactual trajectories generated by rotating the recorded trajectory counterclockwise and clockwise, respectively.
      \textbf{D}: Preference-guided trajectory generated from the annotator's directional preference.
      \textbf{E}: Preference annotation interface, where annotators compare two overlaid trajectories and select the safer or more appropriate alternative. Trajectories are rendered in the robot's egocentric frame; colors distinguish candidates and do not indicate preference.
      }}
      \label{fig:counterfactuals}
      \vspace{-3ex}
\end{figure*}

For each scene, we extract a short-horizon trajectory of fixed length $N$,
\begin{equation}
\tau_0=\{p_1,\dots,p_N\},
\label{eq:traj_base}
\end{equation}
where $p_i=(x_i,y_i,\theta_i)$ denotes the robot's planar pose. For each observation, we construct $M$ candidate trajectories: the recorded trajectory $\tau_0$, two shape-preserving alternatives obtained by rotating $\tau_0$ clockwise and counterclockwise by angles sampled uniformly from a predefined range, and one preference-guided alternative.

To generate the preference-guided alternative, annotators indicate whether the robot should favor the left side, the right side, or neither relative to the recorded trajectory. We randomly select a longer or shorter trajectory toward that side; otherwise, a longer or shorter variant is generated relative to the recorded trajectory. The resulting candidate set is
\begin{equation}
\mathcal{T}_{\mathrm{cf}}
=
\{\tau_0,\tau_1,\dots,\tau_{M-1}\}.
\label{eq:cf_set}
\end{equation}

For preference annotation, annotators compare pairs
$(\tau_i,\tau_j)$ overlaid on the same egocentric observation $x_t$ and select the safer or more appropriate trajectory based on factors such as obstacle clearance, collision risk, nearby agents, and goal progress. This yields a binary preference label
\[
y_{ij} =
\begin{cases}
1 & \text{if $\tau_i$ is preferred over $\tau_j$}, \\
0 & \text{otherwise}.
\end{cases}
\]

For each scene with $M$ counterfactual trajectories, pairwise comparisons induce $\binom{M}{2}$ binary preference annotations due to combinatorial pairing. The resulting preference dataset is
\begin{equation}
\mathcal{P} = \{ (x_t, \tau_i, \tau_j, y_{ij}) \},
\label{eq:pref_dataset}
\end{equation}
where $y_{ij} \in \{0,1\}$ indicates whether trajectory $\tau_i$ is preferred over $\tau_j$ under observation $x_t$. Comparisons labeled as ``both bad'' or ``no preference'' are excluded from
the binary preference dataset and do not contribute to reward learning or
pairwise win counts. The annotation interface, and example counterfactual trajectories used to collect these preferences are shown in Fig.~\ref{fig:counterfactuals}.

Eight annotators participated in the study, with each scene independently
annotated by two annotators. Inter-annotator agreement and Cohen's $\kappa$
are shown in Table~\ref{tab:pref_stats}.


\begin{table}[h]
\centering
\begin{tabular}{l c}
\toprule
Number of annotated observations & 187,920 \\
Counterfactuals per observation ($M$) & 4 \\
Total counterfactual trajectories & 751,680 \\
Fraction where dataset trajectory is not preferred & 70.51\% \\
Total pairwise comparisons & 1,127,520 \\
Inter-annotator agreement & 79.3\% \\
Cohen's $\kappa$ & 0.72 \\
\bottomrule
\end{tabular}
\caption{\small{Statistics of the CHOP preference dataset.}}
\vspace{-10pt}
\label{tab:pref_stats}
\end{table}

\subsection{Trajectory-Conditioned Reward Learning}
\label{sec:reward}

\subsubsection{Trajectory Encoding}

We represent a candidate trajectory as a sequence of ground-plane waypoints
$\tau=(p_1,\ldots,p_K)$, where $p_k=(x_k,y_k)$ and $K$ is determined by the output horizon of the corresponding navigation policy. Each waypoint is mapped to a 128-dimensional embedding using a shared multilayer perceptron. A learned waypoint-index embedding $e_k$ is added to encode its position within the sequence:
\begin{equation}
q_k = f_{\rm wp}(p_k) + e_k.
\label{eq:query}
\end{equation}
The resulting sequence $\{q_k\}_{k=1}^{K}$ provides a learned representation of the candidate trajectory and is used to condition attention over the visual features.
We encode the current RGB observation $I$ using a frozen DINOv3 ViT-S/16 image encoder~\cite{simeoni2025dinov3}, producing a dense visual feature map $F(I)$. The same feature map is shared across all waypoint queries for a candidate trajectory.

\subsubsection{Waypoint-Anchored Deformable Attention}

Inspired by the deformable attention mechanism of Deformable DETR~\cite{zhu2020deformable}, we use each waypoint embedding $q_k$ to retrieve a sparse set of visual features relevant to the candidate trajectory. Unlike object-detection queries, our queries are derived from physical navigation waypoints and are used to construct a trajectory-level preference representation.

For each waypoint query, a learned linear projection predicts a two-dimensional anchor $a_k$ in normalized image coordinates. Each of four attention heads then predicts four sampling offsets and corresponding weights around this anchor. With attention-head index $h$ and sampling index $m$,

\begin{subequations}
\label{eq:attention}
\begin{align}
a_k
&=
\tanh(W_a q_k+b_a),
\label{eq:anchor}\\
\Delta_{khm}
&=
\rho\tanh\!\left((W_\Delta q_k+b_\Delta)_{hm}\right),
\label{eq:offset}\\
\alpha_{khm}
&=
\operatorname{softmax}_{m}
\!\left((W_\alpha q_k+b_\alpha)_{hm}\right),
\label{eq:weights}\\
\tilde{v}_{kh}
&=
\sum_m
\alpha_{khm}
V_h(F(I))
\!\left(a_k+\Delta_{khm}\right),
\label{eq:head_feature}\\
v_k
&=
W_o\operatorname{concat}_{h}
\left[\tilde{v}_{kh}\right]+b_o.
\label{eq:visual_feature}
\end{align}
\end{subequations}

The anchor $a_k\in[-1,1]^2$ is predicted directly from the waypoint embedding rather than obtained from an explicit geometric projection into the image. The scale $\rho$ bounds the maximum normalized sampling offset around each learned anchor. $V_h$ denotes the learned feature projection for attention head $h$, and features at fractional locations are obtained using bilinear sampling. The head-specific features $\tilde{v}_{kh}$ are concatenated and projected to obtain the waypoint-conditioned visual representation $v_k$.

The visual representation is combined with the waypoint embedding as $u_k=q_k+v_k$. A single multi-head transformer encoder layer then models interactions across the waypoint sequence. Mean pooling over the resulting waypoint representations, followed by a layer-normalized MLP, produces the scalar trajectory reward.



\subsubsection{Bradley--Terry Reward Learning}

Given a preferred trajectory $\tau^+$ and a less-preferred trajectory $\tau^-$ under the same image $I$, we model the probability of the observed preference using the Bradley--Terry formulation
\begin{equation}
P_\phi(\tau^+\succ\tau^-\mid I)
=
\sigma\!\left(
R_\phi(I,\tau^+)-R_\phi(I,\tau^-)
\right).
\label{eq:bt}
\end{equation}

The reward model is trained by minimizing the corresponding negative log-likelihood:
\begin{equation}
\mathcal{L}_{\rm BT}(\phi)
=
-\mathbb{E}_{(I,\tau^+,\tau^-)\sim\mathcal{P}}
\log
\sigma\!\left(
R_\phi(I,\tau^+)-R_\phi(I,\tau^-)
\right).
\label{eq:btloss}
\end{equation}

\subsection{Policy Adaptation}
\label{sec:adapt}

Let $o_t$ denote the observation context used by a navigation policy, $g_t$ its native goal input, and $I_t$ the current RGB observation used by the reward model. A pretrained navigation policy predicts
\begin{equation}
\tau_\theta = \pi_\theta(o_t,g_t).
\end{equation}
CHOP-SFT and CHOP-RFT are initialized independently from the corresponding pretrained policy, but use the collected preference supervision differently. For comparison, we construct a standard SCAND-finetuned baseline from the same pretrained checkpoint using the recorded trajectory $\tau_0$ as the supervision target. CHOP-SFT instead replaces $\tau_0$ with the human-preferred trajectory $\tau^\star$.

\subsubsection{CHOP-SFT}

For each annotated observation, the pairwise comparisons over
$\mathcal{T}_{\mathrm{cf}}$ are aggregated to identify a preferred trajectory
$\tau^\star$, defined as the candidate with the largest number of pairwise wins.
If multiple candidates tie, the preference-guided candidate is prioritized,
followed by the recorded trajectory; if the tie remains unresolved, one of the
tied candidates is selected uniformly at random.

For each policy, the preferred trajectory $\tau^\star$ is resampled to match
the policy's native output horizon before computing the supervised loss.
CHOP-SFT then fine-tunes the navigation policy toward $\tau^\star$ using
\begin{equation}
\mathcal{L}_{\mathrm{SFT}}
=
\|\tau_\theta-\tau^\star\|_2^2.
\label{eq:align_loss}
\end{equation}

GNM and ViNT are fine-tuned end-to-end, while OmniVLA is adapted using
LoRA~\cite{hu2022lora}. CHOP-SFT therefore uses the pairwise preferences to
select a supervised trajectory target rather than directly optimizing the
individual preference relations.

\subsubsection{CHOP-RFT}

CHOP-RFT instead uses the learned trajectory-conditioned reward directly for policy adaptation. We freeze the reward model $R_\phi$ and a reference copy $\pi_{\rm ref}$ of the pretrained navigation policy. The reference policy predicts
\begin{equation}
\tau_{\rm ref} = \pi_{\rm ref}(o,g).
\end{equation}
The trainable policy is optimized using
\begin{align}
\mathcal{L}_{\rm RFT}(\theta)
=
\mathbb{E}_{o,g}\big[
&
-R_\phi(I,\tau_\theta)
+
R_\phi(I,\tau_{\rm ref})
\nonumber\\
&
+\beta\,d(\tau_\theta,\tau_{\rm ref})
\big],
\label{eq:rft}\\
d(\tau,\tau')
&=
\frac{1}{2K_\pi}
\sum_{k=1}^{K_\pi}
\|p_k-p'_k\|_2^2.
\label{eq:distance}
\end{align}

Since $R_\phi(I,\tau_{\mathrm{ref}})$ is constant with respect to the policy parameters $\theta$, it does not affect the policy gradient and is included only to express the objective as reward improvement relative to the reference trajectory. Here, $\beta$ controls the strength of the reference-policy regularization.
The reward term encourages the policy to produce trajectories with greater predicted human preference, while the reference penalty constrains deviation from the pretrained policy. The distance $d(\cdot,\cdot)$ is computed over the native $K_\pi$ waypoint outputs of the corresponding navigation policy.

During adaptation, the reward model remains frozen, while gradients with respect to its trajectory input propagate into the navigation policy. CHOP-RFT therefore optimizes the learned preference reward together with reference regularization, without an auxiliary supervised imitation loss. GNM and ViNT are adapted end-to-end, while OmniVLA updates LoRA adapters and its regression/pose heads~\cite{hu2022lora}. The reward model and reference policy are required only during training; deployment uses the adapted navigation policy alone.

\section{Analysis and Results}
\label{sec:experiments}

We evaluate CHOP across both offline navigation benchmarks and real-world
robot deployments. Our experiments address three questions:
\textbf{(i)} whether counterfactual preference supervision improves
safety-aware navigation over standard fine-tuning on the recorded
SCAND trajectories,
\textbf{(ii)} whether the learned trajectory-conditioned reward generalizes to unseen
human preference comparisons, and
\textbf{(iii)} whether retaining the full pairwise preference structure through
CHOP-RFT provides further improvements over winner-based supervised
fine-tuning.

\subsection{Experimental Setup}

We evaluate three state-of-the-art visuomotor navigation policies---OmniVLA~\cite{hirose2025omnivla}, ViNT~\cite{shah2023vint}, and GNM~\cite{shah2022gnm}. For each policy, all variants are initialized from the same pretrained checkpoint. We construct a standard SCAND-finetuned baseline using the recorded trajectory $\tau_0$ as the supervision target. \ours{}-SFT instead fine-tunes the policy using the human-preferred trajectory $\tau^\star$, while \ours{}-RFT performs reward-based adaptation using the learned trajectory-conditioned reward. GNM and ViNT are adapted end-to-end, while OmniVLA is adapted using LoRA~\cite{hu2022lora}.

We use 124 SCAND ROS bags, with 25 held out for testing. The SCAND-finetuned baseline and \ours{}-SFT use the same training observations, differing in their supervision targets, with the former using $\tau_0$ and the latter using $\tau^\star$. The held-out test ROS bags are excluded from all reward learning, policy adaptation, and model selection. Reward-model checkpoints are selected using validation Bradley--Terry loss, and navigation evaluation is performed on the held-out SCAND test set using the same safety and goal-oriented metrics across policy variants.

\subsection{Reward Model Evaluation}
\label{sec:reward_eval}

We first evaluate whether the learned trajectory-conditioned reward generalizes to unseen human preference comparisons and analyze the effect of the reward-model architecture. Table~\ref{tab:reward} reports validation Bradley--Terry (BT) loss and pairwise preference accuracy, where accuracy measures the fraction of comparisons for which the reward assigns a higher score to the human-preferred trajectory.

\begin{table}[t]
\centering\small
\setlength{\tabcolsep}{3pt}
\begin{tabular}{@{}lccrr@{}}
\toprule
Attention & $d/L/H$ & $\lambda_s$ & BT $\downarrow$ & Acc.\% $\uparrow$\\
\midrule
Anchored & 128/1/4 & 0 & \textbf{\RMValBT} & \textbf{\RMValAcc}\\
Anchored & 128/1/4 & .01 & .3322 & 87.17\\
Anchored$^\dagger$ & 128/1/4 & .01 & .3303 & 87.55\\
Anchored & 192/2/6 & .01 & .3310 & 87.32\\
Global & 128/1/4 & 0 & \GlobalValBT & 84.58\\
\bottomrule
\multicolumn{5}{@{}l}{\footnotesize LR $10^{-4}$ except $^\dagger 3\times10^{-5}$. Single runs, no seed intervals.}
\end{tabular}
\caption{Reward-model validation ablations. Each row uses its lowest-validation-BT checkpoint. $d/L/H$ denotes hidden width, path-transformer layers, and attention heads; $\lambda_s$ weights squared-score regularization. The global control matches width/depth, not parameter count or every residual connection.}
\label{tab:reward}
\vspace{-4ex}
\end{table}

The compact anchored-attention model ($d/L/H=128/1/4$) achieves the best overall validation performance, with the lowest BT loss of \RMValBT{} and the highest pairwise accuracy of \RMValAcc\%. We therefore use this configuration for all subsequent \ours{}-RFT experiments.

Adding squared-score regularization ($\lambda_s=0.01$) does not improve performance, and reducing the learning rate partially recovers accuracy but remains below the unregularized model. Increasing the hidden dimension and transformer depth similarly provides no improvement over the compact architecture. The global-attention variant, which allows waypoint queries to attend over the complete image feature map rather than sampling around learned anchors, also performs worse in both BT loss and pairwise accuracy.

On the held-out test set, the selected reward model achieves \RMTestAcc\% pairwise preference accuracy with a Bradley--Terry loss of \RMTestBT. These results show that the learned trajectory reward generalizes to unseen human preference comparisons and provides the reward signal used for \ours{}-RFT policy adaptation.

\subsection{Offline Evaluation on SCAND}
\label{sec:offline_eval}

We evaluate navigation policies offline using recorded observations from the held-out SCAND test set. For each observation, the policy predicts a short-horizon 2D trajectory conditioned on its native visual context and goal representation. Obstacle proximity is measured using synchronized 2D LiDAR scans by computing the minimum Euclidean distance between interpolated predicted trajectory points and observed obstacles. LiDAR is used only for evaluation and is not provided to either the navigation policy or the reward model.

We report three metrics:
\begin{itemize}
    \item \textbf{Near-Collision Count}: Number of predicted trajectories whose minimum obstacle clearance falls below the robot-width.
    \item \textbf{Average Trajectory Deviation (m):} Mean Euclidean distance between corresponding waypoints of the predicted trajectory and the
human-preferred trajectory $\tau^\star$.
    \item \textbf{Average Minimum Obstacle Clearance (m)}: Average minimum distance between the predicted trajectory and observed obstacles.
\end{itemize}

\begin{table}[t]
\centering\small
\setlength{\tabcolsep}{4pt}
\begin{tabular}{@{}llrrr@{}}
\toprule
Policy & Adaptation & NC $\downarrow$ & Dev. $\downarrow$ & Clear. $\uparrow$\\
\midrule
GNM & Base & 5529 & 1.9474 & 2.1629\\
 & \ours-SFT & 2219 & .8056 & 2.9887\\
 & \ours-RFT & \textbf{2156} & \textbf{.8014} & \textbf{3.0329}\\
\midrule
ViNT & Base & 5192 & 1.8082 & 2.2472\\
 & \ours-SFT & 3848 & 1.3216 & 2.5065\\
 & \ours-RFT & \textbf{3497} & \textbf{.8911} & \textbf{2.5972}\\
\midrule
OmniVLA & Base & 2453 & .6345 & 2.7647\\
 & \ours-SFT & 897 & .3185 & 3.0332\\
 & \ours-RFT & \textbf{845} & \textbf{.3143} & \textbf{3.1563}\\ 
\bottomrule
\end{tabular}
\caption{Offline navigation evaluation on the held-out SCAND test set. Near-collision (NC) is reported as a count, while trajectory deviation and obstacle clearance are reported in meters. Lower NC and deviation and higher clearance indicate better performance.}
\label{tab:offline}
\vspace{-5ex}
\end{table}

Table~\ref{tab:offline} compares the SCAND-finetuned baselines with \ours{}-SFT and \ours{}-RFT. \ours{}-SFT substantially improves all three metrics across the evaluated models. Near-collision events decrease by 63.4\%, 25.9\%, and 59.9\% for OmniVLA, ViNT, and GNM, respectively. Preferred-trajectory deviation decreases by 49.8\%, 26.9\%, and 58.6\%, while obstacle clearance increases by 9.7\%, 11.5\%, and 38.2\%. Averaged across the three models, CHOP-SFT reduces near-collision events by 49.7\%, decreases trajectory deviation by 45.1\%, and increases obstacle clearance by 19.8\% relative to the corresponding SCAND-finetuned baselines.

CHOP-RFT further improves performance across all three models. Relative to the corresponding SCAND-finetuned baselines, near-collision events decrease by 65.6\%, 32.6\%, and 61.0\% for OmniVLA, ViNT, and GNM, respectively. Preferred-trajectory deviation decreases by 50.5\%, 50.7\%, and 58.8\%, while obstacle clearance increases by 14.2\%, 15.6\%, and 40.2\%. Averaged across the three models, \ours{}-RFT reduces near-collision events by 53.1\%, decreases trajectory deviation by 53.3\%, and increases obstacle clearance by 23.3\%.

Compared directly with \ours{}-SFT, \ours{}-RFT achieves additional improvements across the same safety and preference-alignment metrics. These results indicate that retaining the pairwise preference structure through reward learning provides a stronger adaptation signal than reducing the preference annotations to a single supervised trajectory target.





\subsection{Real-World Robot Evaluation}
\label{sec:realworld_eval}

To evaluate real-world transfer, we deploy the SCAND-finetuned baseline, CHOP-SFT, and CHOP-RFT policies on a Ghost Robotics Vision60 quadruped in cluttered environments containing both static and dynamic obstacles. All policy variants operate from onboard RGB observations during inference and share the same downstream waypoint-tracking controller. The shared downstream stack performs waypoint tracking only and provides no obstacle avoidance, collision correction, or path replanning. We conduct five trials for every backbone–adaptation–scenario combination, yielding 135 real-world trials in total.

\begin{figure*}[t]
\centering
\includegraphics[width=.9\textwidth]{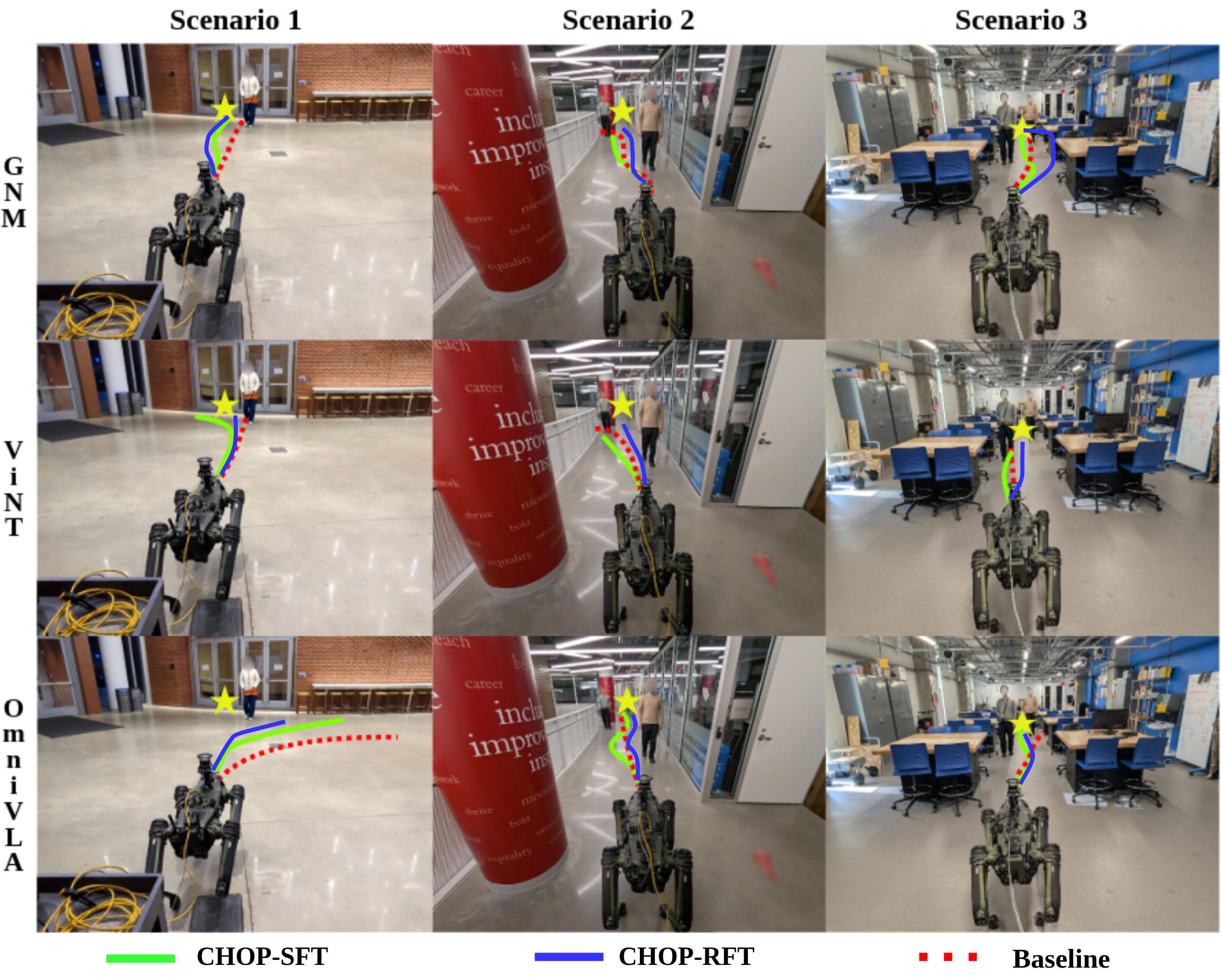}
\caption{\small{
Qualitative comparison of real-world navigation trajectories across three scenarios. Rows correspond to GNM, ViNT, and OmniVLA, and columns correspond to the three evaluation scenarios. For each scene, the predicted trajectories from the SCAND-finetuned baseline, \ours{}-SFT, and \ours{}-RFT policies are overlaid on the robot's egocentric view. The \textbf{red dotted} curve denotes the SCAND-finetuned baseline trained
on $\tau_0$, the \textbf{green} curve denotes \textbf{\ours{}-SFT}, and the \textbf{blue} curve denotes \textbf{\ours{}-RFT}. The \textbf{yellow star} indicates the goal location. In many cases, \ours{}-SFT and \ours{}-RFT produce trajectories with improved obstacle avoidance and better alignment with the goal compared to the baseline.
}}
\vspace{-3ex}
\label{fig:scenarios}
\end{figure*}

We consider three navigation scenarios, illustrated in Fig.~\ref{fig:scenarios}. \textbf{Scenario 1} consists of an open indoor environment where an approaching pedestrian obstructs the direct route to the goal. \textbf{Scenario 2} is a constrained corridor bounded by glass walls and structural pillars, with pedestrians approaching from either side. \textbf{Scenario 3} is a narrow passage between rows of desks in which two pedestrians approach sequentially, requiring repeated obstacle-avoidance maneuvers.

We report:
\begin{itemize}
    \item \textbf{Success Rate (\%):} Percentage of runs that reach within $0.5$\,m of the goal.
    \item \textbf{Average Minimum Obstacle Clearance (m)}: Average minimum distance between the robot and surrounding obstacles along the executed trajectory.
    \item \textbf{Collision/Intervention Events}: Number of physical contacts or safety interventions during each run.
    \item \textbf{Normalized Path Completion}: Fraction of the route completed toward the goal.
\end{itemize}

\begin{table*}[t]
\centering
\small
\setlength{\tabcolsep}{3.2pt}
\begin{tabular}{@{}ll*{12}{c}@{}}
\toprule
& &
\multicolumn{4}{c}{Scenario 1} &
\multicolumn{4}{c}{Scenario 2} &
\multicolumn{4}{c}{Scenario 3} \\
\cmidrule(lr){3-6}
\cmidrule(lr){7-10}
\cmidrule(lr){11-14}
Policy & Adaptation
& S $\uparrow$ & Cl $\uparrow$ & C $\downarrow$ & PC $\uparrow$
& S $\uparrow$ & Cl $\uparrow$ & C $\downarrow$ & PC $\uparrow$
& S $\uparrow$ & Cl $\uparrow$ & C $\downarrow$ & PC $\uparrow$ \\
\midrule

GNM
& Base
& 60 & 4.24 & .40 & .803
& 0 & 1.11 & 1.40 & .360
& 0 & .71 & 1.80 & .412 \\

& \ours{}-SFT
& 100 & 4.57 & .00 & 1.00
& 0 & 1.07 & 1.20 & .463
& 60 & .83 & .40 & .803 \\

& \ours{}-RFT
& 100 & 4.68 & .00 & 1.00
& 20 & 1.26 & 1.10 & .575
& 60 & .91 & .40 & .78 \\

\midrule

ViNT
& Base
& 0 & 4.19 & 1.00 & .359
& 0 & .98 & 1.00 & .625
& 40 & .64 & .80 & .650 \\

& \ours{}-SFT
& 0 & 4.71 & .00 & .888
& 20 & 1.09 & 1.80 & .520
& 40 & .55 & 1.20 & .560 \\

& \ours{}-RFT
& 20 & 4.86 & .00 & .922
& 20 & 1.22 & .00 & .616
& 40 & 1.33 & .00 & .818 \\

\midrule

OmniVLA
& Base
& 0 & 4.88 & .00 & .243
& 80 & 1.02 & .20 & .905
& 0 & .69 & 2.00 & .480 \\

& \ours{}-SFT
& 20 & 4.73 & .20 & .469
& 100 & 1.29 & .00 & 1.000
& 100 & .85 & .00 & 1.000 \\

& \ours{}-RFT
& 40 & 4.89 & .20 & .20
& 100 & 1.31 & .20 & 1.00
& 100 & 0.88 & .00 & 1.00 \\

\bottomrule
\end{tabular}

\caption{
Real-world evaluation across three navigation scenarios.
\textbf{S}: success rate (\%);
\textbf{Cl}: average minimum obstacle clearance (m);
\textbf{C}: mean collision/intervention events per run;
\textbf{PC}: normalized path completion.
}
\label{tab:robot}
\end{table*}

Table~\ref{tab:robot} summarizes the real-world evaluation. Across the three policies, \ours{}-SFT improves average goal success by 28.9 percentage points, increases minimum obstacle clearance by 6.8\%, reduces collision/intervention events by 46.4\%, and improves normalized path completion by 38.7\% relative to the corresponding SCAND-finetuned baselines.

\ours{}-RFT further improves real-world performance, increasing average goal success by 35.6 percentage points and minimum obstacle clearance by 16.0\%, while reducing collision/intervention events by 80.1\% and improving normalized path completion by 42.9\% relative to the corresponding SCAND-finetuned baselines. The reward-based policies therefore retain the safety benefits introduced by counterfactual preference supervision while achieving stronger goal-directed navigation behavior.

\subsection{Discussion}

Across both offline and real-world evaluations, counterfactual human preference supervision consistently improves safety-aware navigation relative to the corresponding SCAND-finetuned policies. \ours{}-SFT demonstrates that replacing the recorded trajectory with a human-preferred counterfactual target substantially improves obstacle avoidance and alignment with human navigation preferences.

\ours{}-RFT provides additional gains by retaining the full pairwise structure of the preference annotations rather than reducing them to a single supervised target. The learned trajectory reward therefore captures preference information that is discarded by winner-based fine-tuning and provides a useful optimization signal beyond the finite set of annotated trajectories.

The magnitude of improvement varies across policy architectures. OmniVLA and GNM benefit most strongly from counterfactual preference supervision, while ViNT exhibits smaller gains, reflecting differences in the underlying capabilities and adaptation behavior of the base policies. 
Nevertheless, the overall trend remains consistent: preference-derived supervision improves navigation relative to the corresponding standard SCAND-finetuned baselines, with CHOP-RFT providing additional improvements over CHOP-SFT.

Together, these results demonstrate two complementary benefits of counterfactual human preferences. \ours{}-SFT improves policy behavior by replacing a single demonstrated action with a human-preferred alternative, while \ours{}-RFT retains the richer pairwise preference structure through a learned trajectory reward and provides further improvements through direct reward-based policy adaptation.

\section{Conclusion}
\ours{} treats alternative local trajectories as objects of human comparison rather than assuming that an executed path is the only appropriate target. We show that counterfactual human preferences provide an effective supervision signal for adapting visuomotor navigation policies through both supervised and reward-based learning. Across GNM, ViNT, and OmniVLA, \ours{}-SFT improves safety and preference alignment relative to standard SCAND fine-tuning on recorded trajectories, while \ours{}-RFT further benefits from retaining the full pairwise preference structure through a learned trajectory-conditioned reward. These improvements transfer to real-world quadruped navigation, demonstrating that counterfactual preference supervision can improve obstacle avoidance and goal-directed behavior beyond standard imitation of recorded demonstrations.

\bibliographystyle{IEEEtran}
\bibliography{references}
\end{document}